\documentclass[a4paper]{article}

\usepackage{INTERSPEECH2019}
\usepackage{graphicx}
\graphicspath{ {./assets/images/} }
\DeclareMathOperator*{\argmax}{arg\,max}
\usepackage{multirow}
\usepackage{verbatim}
\usepackage{tabularx}
\usepackage{algorithm}
\usepackage[noend]{algpseudocode}
\usepackage{subcaption,siunitx,booktabs}
\usepackage{changepage}
\usepackage{contour}
\title{Topical-Chat: Towards Knowledge-Grounded Open-Domain Conversations}
\name{Karthik Gopalakrishnan\thanks{$^*$Equal contribution}$^*$, Behnam Hedayatnia$^*$, Qinlang Chen, Anna Gottardi, Sanjeev Kwatra, Anu Venkatesh, Raefer Gabriel, Dilek Hakkani-T\"ur}

\address{Amazon Alexa AI}
\email{\{karthgop, behnam, qinlangc, gottardi, kwatras, anuvenk, raeferg, hakkanit\}@amazon.com}

\begin{document}

\maketitle
\begin{abstract}
  Building socialbots that can have deep, engaging open-domain conversations with humans is one of the grand challenges of artificial intelligence (AI). To this end, bots need to be able to leverage world knowledge spanning several domains effectively when conversing with humans who have their own world knowledge. Existing knowledge-grounded conversation datasets are primarily stylized with explicit roles for conversation partners. These datasets also do not explore depth or breadth of topical coverage with transitions in conversations. We introduce Topical-Chat, a knowledge-grounded human-human conversation dataset where the underlying knowledge spans 8 broad topics and conversation partners don't have explicitly defined roles, to help further research in open-domain conversational AI. We also train several state-of-the-art encoder-decoder conversational models on Topical-Chat and perform automated and human evaluation for benchmarking.

\end{abstract}

\section{Introduction}

Building conversational bots that can interact with humans in natural language (also known as conversational AI) has been of interest to researchers since the early days of computing, as exemplified by text-based systems such as ELIZA~\cite{weizenbaum1966eliza}. Work on conversational AI generally belongs in one of the following two categories: task-oriented and open-domain. Task-oriented bots aim to help humans accomplish a specific task through multi-turn interactions, whereas open-domain bots aim to serve as social conversation partners with whom humans can have natural and engaging conversations. In addition to mastering traditional language skills like comprehension, open-domain bots (also known as socialbots) need to perfect several conversational skills that come naturally to humans: recalling from world knowledge, reasoning in conjunction with conversational history and constructing valid responses. Socialbots also need to be able to have adequate topical breadth and depth and perform smooth topic transitions.

A critical limiting factor for research into learning these conversational skills is the scarcity of datasets of knowledge-grounded conversations and associated knowledge sources. We introduce Topical-Chat, a dataset of $\sim$11K human-human conversations about knowledge spanning 8 broad topics. Figure~\ref{sample_conversation} depicts a conversation snippet from Topical-Chat, with the full conversation available in Appendix~\ref{appendix:sample_convo}. The dataset was collected by partnering up Amazon Mechanical Turk workers, providing them topical reading sets and asking partners to have naturally coherent and engaging conversations grounded in their provided reading sets. Partners do not have explicitly defined roles they need to serve during a conversation and the reading sets provided to them could be symmetric or asymmetric to varying degrees, which accurately reflects real-world conversations where the world knowledge that both partners gained prior to a conversation may or may not be symmetric. Partners are also asked to annotate each turn of their conversation on several dimensions, such as reading set utilization and sentiment.

In order to create benchmarks for future research with Topical-Chat, we trained several encoder-decoder~\cite{vinyals2015neural,ritter2010unsupervised} conversational models on Topical-Chat, each of which aims to generate a response grounded in a reading set and conditioned on conversational history. We specifically leverage the Transformer architecture \cite{vaswani2017attention} similar to \cite{wow:2019}. We demonstrate the ability of our models to have engaging conversations grounded in knowledge through automated and human evaluation.

\begin{figure}[!tbph]
\begin{tabular}{ l l }
\textbf{Agent}
& \textbf{Message}
\\
\dots & \dots \\
Turker 2 & \begin{tabular}{@{}l@{}}I'd love that job. Visiting Jupiter would be cool \\ too, but that is impossible due to the intense \\ radiation.\end{tabular} \\
\\
Turker 1 & \begin{tabular}{@{}l@{}}Yeah. \textbf{The earth will be helium free by the end} \\ \textbf{of the 21st century.} I wonder if we could make \\ more of it in a lab? Is it even needed?\end{tabular} \\
%Turker 2 & \begin{tabular}{@{}l@{}}No more birthday balloons huh? I wonder if they used helium \\  to make some of those strange creatures talk in Star Wars.\end{tabular} \\
%Turker 1 & \begin{tabular}{@{}l@{}}Could be. I bet we would be surprised in all they did to make \\ movies. Star wars was the first major film to be dubbed into \\ Navajo in 2013. Seems like it should have been done before that\end{tabular} \\
\dots & \dots \\
\end{tabular}
\caption{\label{sample_conversation} {A snippet from a Topical-Chat conversation (sentence used from the corresponding reading set highlighted in bold)}}
\end{figure}

\vspace{-0.5cm}

\section{Related Work}

Recent interest in knowledge-grounded conversations has led to the release of multiple datasets. A dataset of $\sim$4K conversations was released~\cite{zhou2018dataset}, where Wikipedia articles about 30 movies served as the knowledge base. The collection was performed with portions of the articles shown to conversation partners in a scheduled way. A similar dataset of conversations about movies was also released~\cite{moghe:2018}, where the knowledge base comprises Wikipedia articles, reviews and comments mined from the web about $\sim$1K movies. The collection involved self-dialogues, where one crowdworker generates utterances for both sides. More recently, the Wizard of Wikipedia (WoW) dataset~\cite{wow:2019} was released, where the focus, similar to ours, is on collecting open-domain knowledge-grounded conversations. A key difference is their knowledge base comprises Wikipedia articles, whereas we relied on multiple data sources, specifically Washington Post articles and Reddit fun facts in addition to Wikipedia articles about entities, to enable lively interactions.

Sequence-to-sequence generative modeling approaches have become popular for response generation, where the goal is to generate a response given the previous turn in a conversation~\cite{vinyals2015neural,ritter2010unsupervised}. However, responses generated by these sequence-to-sequence models are not always coherent or contextually appropriate and are noted to be often generic and lacking interesting content~\cite{vinyals2015neural}. Such approaches don't explicitly ground responses on relevant knowledge. This has led to work on approaches that include world knowledge into conversational response generation. End-to-end memory networks have been used~\cite{ghazvininejad2018knowledge} to condition the generated responses on knowledge, where attention over the knowledge relevant to the conversation context is estimated, and multiple knowledge representations are included as input during response decoding. Other work~\cite{zhou2018commonsense} retrieves relevant knowledge graphs given the conversation context and encodes the graphs with a static graph attention mechanism. The decoder attentively reads the retrieved knowledge graphs and the knowledge triples within each graph. More recently, a Transformer Memory Network was used~\cite{wow:2019} to encode knowledge sentences and conversation context and decode a response.

\section{Topical-Chat}

Workers on Amazon Mechanical Turk (also known as Turkers) are partnered up and provided highly topical reading sets, and each pair of workers is asked to have a naturally coherent and engaging conversation grounded in their provided reading sets. In our setting, the reading sets provided to conversation partners could be symmetric or have varying degrees of asymmetry, where a pair of reading sets is called symmetric if they contain the exact same information and asymmetric otherwise. This serves as a generalization of the Wizard-Apprentice setting~\cite{wow:2019}. Unlike most (knowledge-grounded or otherwise) conversation settings \cite{wow:2019,weston2016dialog,lewis2017deal,zhang2018personalizing}, the partners do not have explicitly defined roles they need to serve during their conversation. We leverage information asymmetry to implicitly cause both partners to serve dual roles of a \textbf{teacher} and a \textbf{participant} during their conversation. This setting more accurately reflects real-world conversations, where the world knowledge that both partners have gained prior to a conversation may or may not be symmetric. This makes the Topical-Chat dataset\footnotemark versatile, realistic and enables the modeling of both partners.
\footnotetext{Dataset available at \textsf{www.github.com/alexa/Topical-Chat}}

\subsection{Knowledge Base Creation}
To construct reading sets, we created a knowledge base composed of three primitives: \textit{entities}, \textit{facts} and \textit{articles}.

\begin{table}[h]
\caption{\label{topic_entity_budgets} {Topics and their entity budgets}}
\centering
\begin{tabular}{ | l | c | }
\textbf{Topic}
& \textbf{Budget}
\\\hline
Fashion & 20 \\
Politics & 25 \\
Books & 33 \\
Sports & 35 \\
General Entertainment & 38 \\
Music & 39 \\
Science \& Technology & 44 \\
Movies & 66 \\
\hline
Total & 300 \\
\end{tabular}
\end{table}

\textbf{Entity Selection}: We first selected 300 popular entities spanning 8 topics from a prior human-bot conversational dataset collected during a large-scale open-domain socialbot competition between academic research groups~\cite{AlexaPrize:2018}. We specifically selected the entities from all user utterances in this prior dataset, since user utterances inform us what users are interested in talking to socialbots about. To maintain topic diversity, we considered the frequency distribution of the 8 topics across all user utterances to allocate an entity budget $B_i$ for each topic $i$ (with all budgets adding up to 300). We then picked the top-$B_i$ most frequent entities for each topic $i$. The topics and their respective budgets are provided in Table \ref{topic_entity_budgets}.

\textbf{Fact Selection}: We fetched the Wikipedia lead sections of the 300 entities and crowdsourced 8-10 fun facts for each entity using Reddit \cite{reddit}. For each entity, we maintained two versions of the fetched Wikipedia lead sections. The first is a \textit{shortened} version that consists of the first paragraph of the lead section and optionally the second paragraph if the first paragraph contains less than 50 words. The second is a \textit{summarized} version created by extractively summarizing the entire lead section using TextRank \cite{mihalcea2004textrank} into 150 words or less.

\textbf{Article Selection}: We fetched Washington Post articles from 2018 that each referenced 3 or more of the 300 entities and contained 600-1000 words. We removed articles with profane language and then considered the topic-entity budgets to finalize 3088 articles, ensuring adequate coverage for all topics.

\subsection{Reading Sets Creation}
Using the created knowledge base, we construct a pair of reading sets real-time to provide to partners in a conversation. The foundation of a pair of reading sets is an article. For each conversation to be collected, we randomly selected an article from our knowledge base that has not already been used at most 4 times to collect an acceptable conversation. We then apply a random \textit{configuration} from a pre-defined list of configurations to that article. Configurations are defined to impose varying degrees of information symmetry or asymmetry between partners, leading to the collection of a wide variety of conversations.

\subsubsection{Asymmetric Configurations}
%\vspace{-0.4cm}
\begin{figure}[hbt]
  \centering
    \includegraphics[width=\columnwidth]{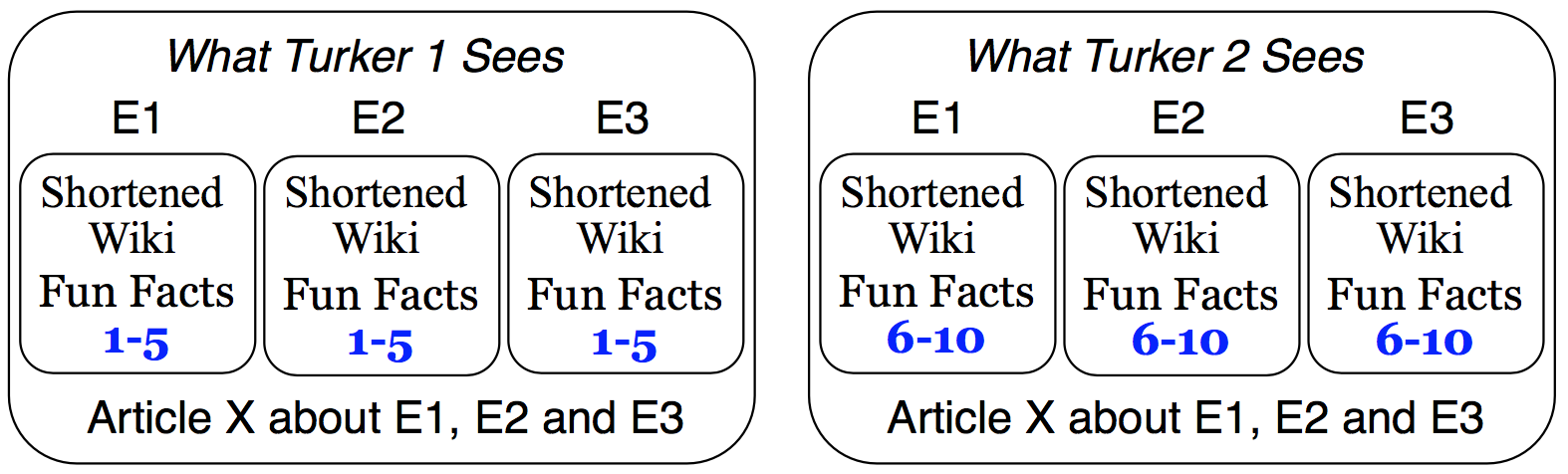}
  \caption{\label{configA}Reading sets for Turkers 1 and 2 in Config A}
\end{figure}

\noindent\textbf{Config A}: Both Turkers get a Washington Post article and shortened Wikipedia lead sections about the top 3 entities by frequency of occurrence in the article. However, they each get a different set of fun facts about these entities. This enables \textbf{asymmetry in entity-level fun facts}.

\begin{figure}[hbt!]
  \centering
    \includegraphics[width=\columnwidth]{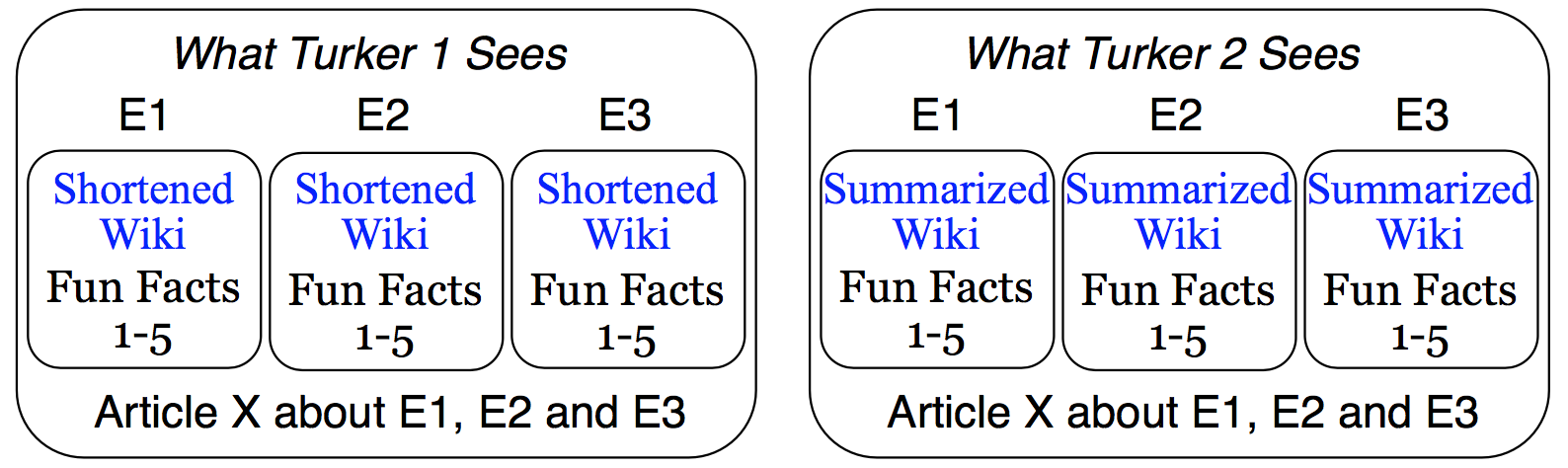}
  \caption{\label{configB}Reading sets for Turkers 1 and 2 in Config B}
\end{figure}

\noindent\textbf{Config B}: Both Turkers get a Washington Post article and 4-5 fun facts about the top 3 entities by frequency of occurrence in the article. However, one Turker gets shortened Wikipedia lead sections and the other gets summarized Wikipedia lead sections about these entities. This enables \textbf{asymmetry in entity-level Wikipedia descriptions}.

\subsubsection{Symmetric Configurations}

\textbf{Config C}: Both Turkers get shortened Wikipedia lead sections and 4-5 fun facts corresponding to the top 3 entities by frequency of occurrence in a Washington Post article. However, the Washington Post article itself is not shown to either Turker. \textbf{Config D}: Both Turkers get a Washington Post article, shortened Wikipedia lead sections and 4-5 fun facts corresponding to the top 3 entities by frequency of occurrence in the article.

\subsection{Conversation Collection}
Qualified workers on Mechanical Turk who take up our Human Intelligence Tasks (also known as HITs) are partnered up and provided topical reading sets to read and consequently chat about. The reading sets are also displayed on the Turkers' screens, near the chat window, during the conversation for reference. All information about an entity \textsf{E1} (shortened/summarized Wikipedia lead sections and fun facts) are displayed as a group titled Factual Section 1. The Washington Post article about entities \textsf{E1}, \textsf{E2} and \textsf{E3} is chunked into 4 similar-sized sections, which are displayed with the titles Article Section 1-4.  Turkers qualify for our HITs if their past approved HITs and approval rates are at least 1000 and 99\% respectively, ensuring our conversations involve experienced Turkers. We used a customized version of the ParlAI \cite{miller2017parlai} framework to collect conversations. We allow partner Turkers to submit their conversation only if they have conversed for at least 20 turns. At each turn during a conversation, while they are waiting for their partner to respond, we ask each partner to: annotate the sentiment of their message on an 8-point scale (Angry, Disgusted, Fearful, Sad, Happy, Surprised, Curious to Dive Deeper, Neutral), specify the knowledge source used to generate their message (Factual Section 1-3, Article Section 1-4 and/or Personal Knowledge) and rate the quality of their partner's previous message on a 5-point scale (Poor, Not Good, Passable, Good and Excellent). At the end of a conversation, we ask both partners to rate the quality of the conversation on the same 5-point scale.

We relied on a mix of manual reviewing and automated checks to ensure the conversations we were collecting were acceptable. The automated checks involved computing and verifying our quality metrics were above tuned thresholds. Turkers who had conversations of exceptionally high quality were awarded bonuses. Some statistics about our dataset are shown in Table~\ref{topical_chat_stats}. We created two versions of the validation and test set: \textit{frequent} and \textit{rare}. The frequent set contains entities frequently seen in the training set, while the rare set contains entities that were infrequently or never seen in the training set. The presence of multiple entities per conversation by design made it harder to perform a perfect entity-level split of our dataset unlike in~\cite{wow:2019}, where this is much easier to accomplish since each conversation is associated with a single entity (referred to as a topic in their paper). Appendix~\ref{appendix:val_test_split} describes our approach.

\begin{table*}[t]
%\begin{minipage}{.5\linewidth}
\centering
\caption{\label{topical_chat_stats} {Topical-Chat conversation stats}}
\scalebox{0.9}{
\begin{tabular}{ c|c|c|c|c|c|c }
\hline
\textbf{Topical-Chat} & \textbf{Config} & \textbf{Train} & \textbf{Valid Freq.} & \textbf{Valid Rare} & \textbf{Test Freq.} & \textbf{Test Rare} \\ \hline
\multirow{6}{*}{Number of Conversations} & A & 2199 & 141 & 127 & 131 & 136 \\
 & B & 2114 & 144 & 138 & 141 & 154 \\
 & C & 2259 & 150 & 143 & 125 & 139 \\
 & D & 2486 & 130 & 158 & 168 & 136 \\ \cline{2-7}
 & \textbf{Total} & 9058 & 565 & 566 & 565 & 565 \\ \hline
 \multirow{6}{*}{Number of Utterances} & A & 48022 & 3083 & 2792 & 2875 & 2955  \\
  & B & 46098 & 3177 & 3066 & 3116 & 3348  \\
  & C & 49705 & 3248 & 3237 & 2737 & 3012 \\
  & D & 54481 & 2859 & 3445 & 3735 & 3023 \\ \cline{2-7}
  & \textbf{Total} & 198306 & 12367 & 12540 & 12463 & 12338 \\ \hline
  \multirow{6}{*}{Average Number of Turns per Conversation} & A & 21.8 & 21.8 & 22.0 & 21.9 & 21.7 \\
   & B & 21.8 & 22.0 & 22.2 & 22.1 & 21.7 \\
   & C & 22.0 & 21.6 & 22.6 & 21.9 & 21.7 \\
   & D & 21.9 & 22.0 & 21.8 & 22.2 & 22.2\\ \cline{2-7}
   & \textbf{Total} & 21.9 & 21.9 & 22.1 & 22.0 & 21.8 \\ \hline
   \multirow{6}{*}{Average Length of Utterance} & A & 19.7 & 19.9 & 20.2 & 19.4 & 19.4  \\
    & B & 19.7 & 20.1 & 19.0 & 19.1 & 20.2 \\
    & C & 19.6 & 20.1 & 19.1 & 20.0 & 19.9 \\
    & D & 19.7 & 19.2 & 19.6 & 20.0 & 20.0 \\ \cline{2-7}
    & \textbf{Total} & 19.7 & 19.8 & 19.8 & 19.6 & 19.9\\ \hline
\end{tabular}}
%\end{minipage}
\end{table*}
%\begin{table}

\section{Models}
Let us denote a partial conversation $C_j = [x_1,\dots, x_j]$, where for $1 \leq i \leq j$, $x_i$ is the $i^{th}$ turn in the conversation. Our conversation history is denoted as $H_j = x_1 \oplus \dots \oplus x_j$, which is a flattened sequence of all tokens in $C_j$. $x_{j+1}$, the ground-truth response at turn $j+1$, is our target sequence to be predicted for all models. Denote the reading set corresponding to the Turker associated with turn $j+1$ as $R$, which we tokenize into a series of knowledge candidate sentences [$k_i$], $i = 1,\dots, N_R$. Denote $W_K$ as a truncate parameter for a knowledge sentence $K$, which retains at most $W_K$ tokens from the start in $K$. Denote $W_H$ as a truncate parameter for a conversation history $H$, which retains at most $W_H$ tokens from the end in $H$.

\subsection{Transformer}
We train a Transformer with ($H_j, x_{j+1}$) pairs. During inference, it decodes a response $y$ given a conversation history $H$.

\subsection{Transformer with Knowledge}

$H_j$ and a selected sentence $\hat{k}$ from [$k_i$] are encoded with a shared Transformer, concatenated and passed to the Transformer decoder. Knowledge selection in the absence of ground-truth response $x_{j+1}$ is an open problem. We currently utilize $x_{j+1}$ in the argmax oracle to select $\hat{k}$, as follows:
\begin{equation*}
    \hat{k} = \argmax_{[k_i]} \left( \frac{\vec{x_{j+1}} \cdot \vec{k_i}}{\left\lVert \vec{x_{j+1}} \right\rVert \left\lVert \vec{k_i} \right\rVert} \right)
\end{equation*}

$\vec{x_{j+1}}$ and $\vec{k_i}$ are TF-IDF vectors for $x_{j+1}$ and $k_i$. The TF-IDF vectorizer is learned by sentence-tokenizing all reading sets in Topical-Chat and treating each sentence as a document.

\begin{figure}[hbt]
  \centering
    \includegraphics[width=\columnwidth]{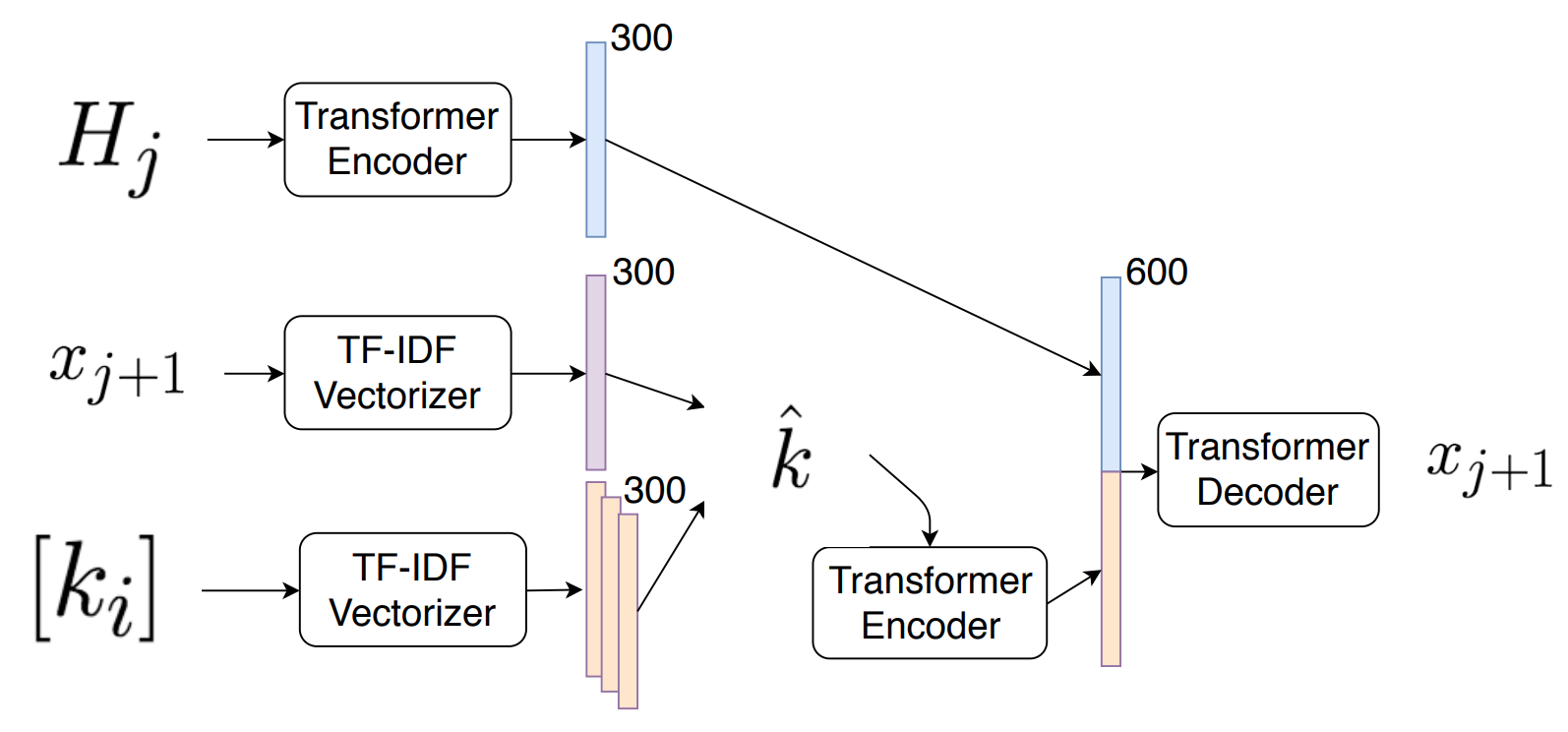}
  \caption{Transformer with knowledge}
  \label{fig:transformerknowledge}
\end{figure}

\vspace{-0.6cm}
\section{Experiments}
All models were trained using ParlAI~\cite{miller2017parlai}. Our Transformer contains two layers with two attention heads and a feed-forward hidden layer size of 300 with dropout 0.2. We randomly initialized 300-dimensional word embeddings, which are learned during training. We do not learn positional embeddings and encode position using one-hot vectors. We use a batch size of 32, stochastic gradient descent for optimization with a gradient clip of 0.1 and learning rate scheduler decay 0.5 with patience 3. We stop training when perplexity on the validation frequent set does not decrease for 10 epochs. We use beam search with a beam size of 5 for decoding.

We also experimented with pre-training the Transformer on BookCorpus~\cite{bookcorpusdataset} using a language modeling objective of maximizing the log-likelihood of the next token given a context window of tokens~\cite{radford2018improving}.
We use byte-pair encoding (BPE)~\cite{sennrich2015neural} when pre-training (vocabulary size 37758). When not pre-training, we do not use BPE (vocabulary size 49957).

\vspace{-0.1cm}
\section{Results}
We use the following acronyms for models for the sake of brevity: TF = Transformer, w/ p.t. = with pre-training, w/ k. = with knowledge. We used a large $W_K$ = 128 when using knowledge, effectively making the parameter irrelevant in our setting since most knowledge sentences have fewer than 128 tokens. In order to decide on an appropriate $W_H$, we tried training a Transformer that uses knowledge with varying $W_H$ and evaluated them on automated metrics described below (Table~\ref{tab:topical_chat_automatic_results_beam}). We observe that $W_H$ = 32 works best. We believe this reflects our knowledge model's inability to attend to important tokens in the dialog context when a large $W_H$ is used. Consequently, we used $W_H$ = 32 in Tables~\ref{tab:topical_chat_automatic_results} and~\ref{tab:topical_chat_human_results}.

For automated evaluation, we consider metrics such as perplexity (PPL), unigram F1 of model prediction with ground-truth response and $n$-gram diversity (Div.)~\cite{ghazvininejad2018knowledge}. In Table~\ref{tab:topical_chat_automatic_results}, we observe that all our models have high unigram and bigram diversity, demonstrating that the models learn to decode responses that are lexically informative and diverse. We also observe an improvement in unigram F1 and increase in PPL when knowledge is used.

We performed human evaluation of our models by first creating 150 evaluation snippets, each comprising \{$C_j$, $\hat{k}$, [$r_c$]\}, $c$ = $1 \dots N$, where [$r_c$] is a set of $N$ responses ($N-1$ from trained models and one ground-truth response $x_{j+1}$) given a partial conversation $C_j$ and selected sentence $\hat{k}$. The partial conversation corresponding to each snippet came from a distinct conversation in the Topical-Chat test frequent set. For each $r_c$ in each snippet, we asked two humans to separately annotate~\cite{venkatesh2018evaluating,see2019makes} (possible values in parentheses) whether $r_c$ is comprehensible (0/1), on-topic (0/1) and interesting (0/1). We also asked them to annotate how effectively $\hat{k}$ is utilized in $r_c$ (0-3) and if they would have liked to continue the conversation after $r_c$ (0/1). We computed Cohen's kappa for binary and Fleiss' kappa for nominal-scale annotations as measures of reliability of agreement and observed poor agreement for interesting (0.29) and continue conversation (0.27). Consequently, we aggregate and report mean annotation scores for parameters with high agreement in Table~\ref{tab:topical_chat_human_results}. We use the following acronyms for the sake of brevity: comprehensible = comp., on-topic = o.t., leverage knowledge = l.k. We observe that all models are rated to mostly produce comprehensible responses and the models that ingest knowledge are rated to produce responses that leverage them, albeit only somewhat effectively.

%\vspace{-0.3cm}
\begin{table}[!tbph]\centering
\caption{Automated metrics on test set (Frequent / Rare)} \label{tab:topical_chat_automatic_results}
\scalebox{0.9}{\begin{tabular}{l | c | c | c | c }
Model
%& \multicolumn{4}{c|}{Seen/Unseen }\\\hline
       & PPL & F1 & Div. ($n$=1) & Div. ($n$=2) \\\hline
%TF & 42.3/44.5 & 0.01/0.01 & 0.86/0.87 & 0.86/0.87 \\
TF & 29.8 / 40.4 & 0.16 / 0.16 & 0.85 / 0.84 & \textbf{0.86 / 0.86} \\
%TF (w/ p.t.) & 39.3/41.5 & 0.01/0.01 & 0.86/0.85 & 0.86/0.86 \\
TF (w/ p.t.) & 26.3 / 36.3 & 0.16 / 0.15 & \textbf{0.86 / 0.85} & 0.86 / 0.85  \\
%TF (w/ k.) & 44.8/46.5 & 0.02/0.02 & 0.83/0.83 & 0.81/0.81 \\
TF (w/ k.) & 33.8 / 43.6 & \textbf{0.22 / 0.20} & 0.84 / 0.80 & 0.83 / 0.81 \\
%TF (w/ k. p.t.) & 36.5/38.2 & 0.02/0.02 & 0.81/0.81 & 0.82/0.82\\
TF (w/ k. p.t.) & 34.1 / 44.8 & 0.22 / 0.19 & 0.85 / 0.82 & 0.84 / 0.82 \\
\end{tabular}}
\end{table}

%\contourlength{0.1pt}
%\contournumber{10}
\vspace{-0.6cm}
\begin{table}[!tbph]\centering
\caption{Human evaluation metrics for 150 test freq. snippets} \label{tab:topical_chat_human_results}
\scalebox{0.9}{\begin{tabular}{l | c | c | c}
Model
\footnotemark
%& \multicolumn{5}{c|}{Annotators 1/2}\\\hline
       & comp. ($\kappa$ = 0.83) & o.t. ($\kappa$ = 0.67) & l.k. ($\kappa$ = 0.62) \\\hline
Human & \textbf{0.99} & \textbf{0.93} & \textbf{1.92} \\
TF & 0.87 & 0.60 & 0.08\\
TF (w/ p.t.) & \underline{0.88} & 0.62 & 0.12\\
TF (w/ k.) & 0.78 & \underline{0.69} & 0.63 \\
TF (w/ k. p.t.) & 0.71 & 0.66 & \underline{0.80}\\
\end{tabular}}
\end{table}
\footnotetext{Models used for human evaluation were trained on a subset of the training set.}

\vspace{-0.6cm}
\begin{table}[!tbph]\centering
\caption{Effect of varying $W_H$ for TF (w/ k.) on test freq.} \label{tab:topical_chat_automatic_results_beam}
\scalebox{0.9}{\begin{tabular}{l | c | c | c | c }
$W_H$
& PPL
& F1 & Div. ($n$=1) & Div. ($n$=2) \\\hline
%16 & 35.4 & 0.02 & 0.66 & 0.68 \\
16 & 34.2 & 0.19 & 0.68 & 0.70 \\
%32 & 34.8 & 0.03 & 0.86 & 0.85  \\
32 & \textbf{33.8} & \textbf{0.22} & \textbf{0.84} & \textbf{0.83} \\
%64 & 37.7 & 0.02 & 0.85 & 0.82 \\
64 & 42.5 & 0.21 & 0.81 & 0.81 \\
%128 & 39.7 & 0.02 & 0.83 & 0.82 \\
128 & 38.3 & 0.21 & 0.80 & 0.80 \\
%512 & 44.8 & 0.02 & 0.83 & 0.81 \\
512 & 42.7 & 0.19 & 0.82 & 0.81 \\
\end{tabular}}
\end{table}

\vspace{-0.6cm}
\section{Conclusion}
We introduce Topical-Chat, an open-domain knowledge-grounded conversation dataset without explicit roles for conversation partners and containing depth and breadth of topical coverage with transitions in conversations. We train simple Transformer-based models for response generation and evaluate them using automated metrics for benchmarking. We also provide evidence of qualitative value through human evaluation of these models. We hope that the release of Topical-Chat fosters data-driven research in open-domain knowledge-grounded conversational AI.

\bibliographystyle{IEEEtran}

\bibliography{mybib}

\begin{thebibliography}{10}
\providecommand{\url}[1]{#1}
\csname url@samestyle\endcsname
\providecommand{\newblock}{\relax}
\providecommand{\bibinfo}[2]{#2}
\providecommand{\BIBentrySTDinterwordspacing}{\spaceskip=0pt\relax}
\providecommand{\BIBentryALTinterwordstretchfactor}{4}
\providecommand{\BIBentryALTinterwordspacing}{\spaceskip=\fontdimen2\font plus
\BIBentryALTinterwordstretchfactor\fontdimen3\font minus
  \fontdimen4\font\relax}
\providecommand{\BIBforeignlanguage}[2]{{%
\expandafter\ifx\csname l@#1\endcsname\relax
\typeout{** WARNING: IEEEtran.bst: No hyphenation pattern has been}%
\typeout{** loaded for the language `#1'. Using the pattern for}%
\typeout{** the default language instead.}%
\else
\language=\csname l@#1\endcsname
\fi
#2}}
\providecommand{\BIBdecl}{\relax}
\BIBdecl

\bibitem{weizenbaum1966eliza}
J.~Weizenbaum \emph{et~al.}, ``Eliza---a computer program for the study of
  natural language communication between man and machine,''
  \emph{Communications of the ACM}, vol.~9, no.~1, pp. 36--45, 1966.

\bibitem{vinyals2015neural}
O.~Vinyals and Q.~Le, ``A neural conversational model,'' \emph{arXiv preprint
  arXiv:1506.05869}, 2015.

\bibitem{ritter2010unsupervised}
A.~Ritter, C.~Cherry, and B.~Dolan, ``Unsupervised modeling of twitter
  conversations,'' in \emph{Human Language Technologies: The 2010 Annual
  Conference of the North American Chapter of the Association for Computational
  Linguistics}.\hskip 1em plus 0.5em minus 0.4em\relax Association for
  Computational Linguistics, 2010, pp. 172--180.

\bibitem{vaswani2017attention}
A.~Vaswani, N.~Shazeer, N.~Parmar, J.~Uszkoreit, L.~Jones, A.~N. Gomez,
  {\L}.~Kaiser, and I.~Polosukhin, ``Attention is all you need,'' in
  \emph{Advances in Neural Information Processing Systems}, 2017, pp.
  5998--6008.

\bibitem{wow:2019}
E.~Dinan, S.~Roller, K.~Shuster, A.~Fan, M.~Auli, and J.~Weston, ``Wizard of
  wikipedia: Knowledge-powered conversational agents,'' \emph{arXiv preprint
  arXiv:1811.01241}, 2018.

\bibitem{zhou2018dataset}
K.~Zhou, S.~Prabhumoye, and A.~W. Black, ``A dataset for document grounded
  conversations,'' \emph{arXiv preprint arXiv:1809.07358}, 2018.

\bibitem{moghe:2018}
N.~Moghe, S.~Arora, S.~Banerjee, and M.~M. Khapra, ``Towards exploiting
  background knowledge for building conversation systems,'' 2018.

\bibitem{ghazvininejad2018knowledge}
M.~Ghazvininejad, C.~Brockett, M.-W. Chang, B.~Dolan, J.~Gao, W.-t. Yih, and
  M.~Galley, ``A knowledge-grounded neural conversation model,'' in
  \emph{Thirty-Second AAAI Conference on Artificial Intelligence}, 2018.

\bibitem{zhou2018commonsense}
H.~Zhou, T.~Young, M.~Huang, H.~Zhao, J.~Xu, and X.~Zhu, ``Commonsense
  knowledge aware conversation generation with graph attention.'' in
  \emph{IJCAI}, 2018, pp. 4623--4629.

\bibitem{weston2016dialog}
J.~E. Weston, ``Dialog-based language learning,'' in \emph{Advances in Neural
  Information Processing Systems}, 2016, pp. 829--837.

\bibitem{lewis2017deal}
M.~Lewis, D.~Yarats, Y.~N. Dauphin, D.~Parikh, and D.~Batra, ``Deal or no deal?
  end-to-end learning for negotiation dialogues,'' \emph{arXiv preprint
  arXiv:1706.05125}, 2017.

\bibitem{zhang2018personalizing}
S.~Zhang, E.~Dinan, J.~Urbanek, A.~Szlam, D.~Kiela, and J.~Weston,
  ``Personalizing dialogue agents: I have a dog, do you have pets too?'' in
  \emph{Proceedings of the 56th Annual Meeting of the Association for
  Computational Linguistics (Volume 1: Long Papers)}, 2018, pp. 2204--2213.

\bibitem{AlexaPrize:2018}
C.~Khatri, B.~Hedayatnia, A.~Venkatesh, J.~Nunn, Y.~Pan, Q.~Liu, H.~Song,
  A.~Gottardi, S.~Kwatra, S.~Pancholi, M.~Cheng, Q.~Chen, L.~Stubel,
  K.~Gopalakrishnan, K.~Bland, R.~Gabriel, A.~Mandal, D.~Hakkani-T\"ur,
  G.~Hwang, N.~Michel, E.~King, and R.~Prasad, ``Advancing the state of the art
  in open domain dialog systems through the alexa prize,'' in \emph{Alexa Prize
  Proceeedings {\rm
  (https://developer.amazon.com/alexaprize/challenges/past-challenges/2018/)}},
  2018.

\bibitem{reddit}
Reddit, ``r/todayilearned,'' \url{https://www.reddit.com/r/todayilearned/}.

\bibitem{mihalcea2004textrank}
R.~Mihalcea and P.~Tarau, ``Textrank: Bringing order into text,'' in
  \emph{Proceedings of the 2004 conference on empirical methods in natural
  language processing}, 2004.

\bibitem{miller2017parlai}
A.~H. {Miller}, W.~{Feng}, A.~{Fisch}, J.~{Lu}, D.~{Batra}, A.~{Bordes},
  D.~{Parikh}, and J.~{Weston}, ``Parlai: A dialog research software
  platform,'' \emph{arXiv preprint arXiv:{1705.06476}}, 2017.

\bibitem{bookcorpusdataset}
BookCorpus, \url{https://github.com/soskek/bookcorpus/}.

\bibitem{radford2018improving}
A.~Radford, K.~Narasimhan, T.~Salimans, and I.~Sutskever, ``Improving language
  understanding by generative pre-training,'' \emph{URL
  https://s3-us-west-2.amazonaws.com/openai-assets/research-covers/language-unsupervised/language\_understanding\_paper.pdf},
  2018.

\bibitem{sennrich2015neural}
R.~Sennrich, B.~Haddow, and A.~Birch, ``Neural machine translation of rare
  words with subword units,'' \emph{arXiv preprint arXiv:1508.07909}, 2015.

\bibitem{venkatesh2018evaluating}
A.~Venkatesh, C.~Khatri, A.~Ram, F.~Guo, R.~Gabriel, A.~Nagar, R.~Prasad,
  M.~Cheng, B.~Hedayatnia, A.~Metallinou, R.~Goel, S.~Yang, and A.~Raju, ``On
  evaluating and comparing open domain dialog systems,'' 2018.

\bibitem{see2019makes}
A.~See, S.~Roller, D.~Kiela, and J.~Weston, ``What makes a good conversation?
  how controllable attributes affect human judgments,'' 2019.

\end{thebibliography}

\appendix

\section{Valid / Test Set Creation Strategy}
\label{appendix:val_test_split}

We adopt a greedy strategy for creating the validation and test splits from the collected data. We first create a list of all entity triplets for all collected conversations. We define entity frequency as the number of triplets containing an entity and consequently assign a score for each triplet as the sum of its constituent entity frequencies. We sort the list of all triplets in increasing order of their scores, extract the top 10\% of the triplets and create two equal or off-by-one sized partitions $V_{rare}$ (validation rare) and $T_{rare}$ (test rare) of the corresponding conversations. Next, we randomly select 80\% of the triplets and create partition $D_{train}$ (train) of the corresponding conversations. Finally, we create two equal or off-by-one sized partitions $V_{freq}$ (validation frequent) and $T_{freq}$ (test frequent) of the remaining conversations.

\newpage
\onecolumn

\section{Topical-Chat: Sample Conversation}
\label{appendix:sample_convo}

\begin{table}[H]
\caption{\label{sample_conversation_appendix} {The reading sets corresponding to this conversation were constructed according to Config A, where the asymmetry lies in entity-level fun facts. We omit the reading sets from the paper in the interest of brevity, but for reference, the article was headlined ``\textit{May the 4th be with Earth, a perfect planet that's not too Hoth or Tatooine}'' and the entities were \textit{Earth}, \textit{Planet} and \textit{Star Wars}.}}
\begin{tabularx}{\textwidth}{ | l | X | }
\hline
\textbf{Agent}
& \textbf{Message}
\\\hline
Turker 1 & Hi, how are you? \\
Turker 2 & Good, just happy to be living on planet Earth and not on part of the 71\% that is covered with water. \\
Turker 1 & Me too! Like Water World. I do wish we had that 80\% of forests that are now gone. \\
Turker 2 & Me too. But its hard for trees to grow in some areas, like the polar regions that are covered in ice. \\
Turker 1 & Agreed. When the earth was first formed a day was only 5.5 hours long. \\
Turker 2 & That's pretty neat. That would probably have made the tides crazy that are caused by the gravitational interaction between the Earth and the Moon. \\
Turker 1 & The interior is made of solid iron core, a liquid otter core that generates the earth's magnetic field. \\
Turker 2 & I did not know that. I did know that Jupiter saves the Earth from many asteroid impacts every year. \\
Turker 1 & That Jupiter is a standup planet. Always there when you need someone to block asteroids... Also there in case of an alien invasion, the UN's ambassador to extraterrestrials! \\
Turker 2 & I'd love that job. Visiting Jupiter would be cool too, but that is impossible due to the intense radiation. \\
Turker 1 & Yeah. The earth will be helium free by the end of the 21st century. I wonder if we could make more of it in a lab? Is it even needed? \\
Turker 2 & No more birthday balloons huh? I wonder if they used helium to make some of those strange creatures talk in Star Wars. \\
Turker 1 & Could be. I bet we would be surprised in all they did to make movies. Star wars was the first major film to be dubbed into Navajo in 2013. Seems like it should have been done before that... \\
Turker 2 & Yeah, you would think. Apparently the entire saga is told from the perspective of R2D2. I never knew that. \\
Turker 1 & Really? I didn't know that. I didn't know Poe was Guatemalan either. \\
Turker 2 & Huh, interesting trivia fact. I certainly didn't know that. Another interesting fact is that the Star Wars franchise is worth an estimated \$65 billion. That's a nice chunk of change. \\
Turker 1 & Right? James earl jones only spent 2.5 hours recording darth vadars lines for star wars. \\
Turker 2 & I had no idea. I forgot he wasn't actually in the costume. Star Wars is truly an American epic, and Darth Vader is iconic. \\
Turker 1 & I agree! He did it for only \$7500 too!!! \\
Turker 2 & What a bargain! Well, it's been nice chatting with you. \\
Turker 1 & Nice chatting with you! \\
\hline
\end{tabularx}
\end{table}

\end{document}